\documentclass[lettersize,journal]{IEEEtran}
\usepackage{amsmath, amsfonts}
\usepackage{array}
\usepackage{textcomp}
\usepackage{url}
\usepackage{verbatim}
\usepackage{graphicx}
\usepackage{tcolorbox}
\usepackage{cite}
\usepackage{lipsum}
\usepackage{multicol}
\usepackage{colortbl}
\usepackage{algpseudocode}
\usepackage{booktabs}
\usepackage{listings}
\usepackage{balance}
\usepackage{hyperref}
\usepackage{subcaption}
\usepackage{float}
\usepackage{dblfloatfix}
\usepackage{soul}
\usepackage{makecell}
\definecolor{lightgray}{rgb}{0.95, 0.95, 0.95}
\definecolor{purple}{rgb}{0.58, 0, 0.82}

\tcbset{
    myboxstyle/.style={
        colback=blue!5!white, 
        colframe=black!75,    
        fonttitle=\bfseries,  
        coltitle=white,       
        boxrule=0.75pt,       
        arc=6pt,              
        outer arc=6pt,
        width=\columnwidth,   
        left=6pt,             
        right=6pt,            
        top=6pt,              
        bottom=6pt,            
        before skip=10pt,    
        after skip=10pt,     
    }
}
\newtcolorbox[auto counter, list inside=pabox]{pabox}[2][]{%
colback= blue!5!white,
colframe=black!75, 
fonttitle=\bfseries,
coltitle=white,
boxrule=0.75pt,       
arc=6pt,              
outer arc=6pt,
width=\columnwidth,   
left=6pt,             
right=6pt,            
top=6pt,              
bottom=2pt,            
before skip=5pt,    
after skip=1pt,     
title=Box~\thetcbcounter: #2,#1}

\def\BibTeX{{\rm B\kern-.05em{\sc i\kern-.025em b}\kern-.08em
    T\kern-.1667em\lower.7ex\hbox{E}\kern-.125emX}}

\definecolor{SentimentBlue}{HTML}{67ade0}
\definecolor{SentimentRed}{HTML}{fc9b9a}
\definecolor{SentimentGreen}{HTML}{86e393}

\begin{document}
\title{Assessing Ethical Reasoning in Efficient Multiplexity-Infused Language Models}
\title{Auditing the Worldview of Large Language Models: An Applied Multiplexity Perspective}
\title{Rethinking AI in Education: Auditing Large Language Models Through a Multiplexity Lens}
\title{Towards Pluralistic Educational AI for the World: Evaluating Frontier Large Language Models for Cultural Biases through Applied Multiplexity}
\title{Towards Pluralistic Educational AI for the World: Auditing Frontier LLMs Through An Applied Multiplexity Perspective}
\title{Toward Inclusive Educational AI: Auditing Frontier LLMs through a Multiplexity Lens}

\author{
    Abdullah Mushtaq\IEEEauthorrefmark{1}, 
    Muhammad Rafay Naeem\IEEEauthorrefmark{1}, 
    Muhammad Imran Taj\IEEEauthorrefmark{2}, 
    Ibrahim Ghaznavi\IEEEauthorrefmark{1}, 
    Junaid Qadir\IEEEauthorrefmark{3} \\

    \IEEEauthorblockA{\IEEEauthorrefmark{1}Department of Computer Science, Information Technology University, Lahore, Pakistan \\
    \{bscs20078, bscs20004, ibrahim.ghaznavi\}@itu.edu.pk} \\

    \IEEEauthorblockA{\IEEEauthorrefmark{2}College of Interdisciplinary Studies, Zayed University, Dubai, UAE \\
    MuhammadImran.Taj@zu.ac.ae} \\

    \IEEEauthorblockA{\IEEEauthorrefmark{3}Department of Computer Science and Engineering, Qatar University, Doha, Qatar \\
    jqadir@qu.edu.qa}


}

\definecolor{customYellow}{RGB}{236,234,223} 
\maketitle

\begin{abstract}
As large language models (LLMs) like GPT-4 and Llama 3 become integral to educational contexts, concerns are mounting over the cultural biases, power imbalances, and ethical limitations embedded within these technologies. Though generative AI tools aim to enhance learning experiences, they often reflect values rooted in Western, Educated, Industrialized, Rich, and Democratic (WEIRD) cultural paradigms, potentially sidelining diverse global perspectives. This paper proposes a framework to assess and mitigate cultural bias within LLMs through the lens of applied multiplexity. Multiplexity, inspired by Senturk et al. and rooted in Islamic and other wisdom traditions, emphasizes the coexistence of diverse cultural viewpoints, supporting a multi-layered epistemology that integrates both empirical sciences and normative values. Our analysis reveals that LLMs frequently exhibit cultural polarization, with biases appearing in both overt responses and subtle contextual cues. To address inherent biases and incorporate multiplexity in LLMs, we propose two strategies: \textit{Contextually-Implemented Multiplex LLMs}, which embed multiplex principles directly into the system prompt, influencing LLM outputs at a foundational level and independent of individual prompts, and \textit{Multi-Agent System (MAS)-Implemented Multiplex LLMs}, where multiple LLM agents, each representing distinct cultural viewpoints, collaboratively generate a balanced, synthesized response. Our findings demonstrate that as mitigation strategies evolve from contextual prompting to MAS-implementation, cultural inclusivity markedly improves, evidenced by a significant rise in the Perspectives Distribution Score (PDS) and a PDS Entropy increase from 3.25\% at baseline to 98\% with the MAS-Implemented Multiplex LLMs. Sentiment analysis further shows a shift towards positive sentiment across cultures, with the MAS-Implemented Multiplex LLMs achieving 0\% negative sentiment, underscoring enhanced cultural sensitivity. This pioneering study establishes a baseline for assessing and fostering cultural inclusivity in educational AI, laying the groundwork for a globally pluralistic approach that respects diverse cultural perspectives. We hope this work inspires further research toward creating AI technologies that serve a truly inclusive and multicultural educational ecosystem.
\end{abstract}

\begin{IEEEkeywords}
LLMs, Cultural Bias, Ethics, Education, Pluralistic AI, Multiplexity
\end{IEEEkeywords}

\section{Introduction} \label{sec:Introduction}

\IEEEPARstart{T}{he} rapid advancement of Large Language Models (LLMs) has significantly impacted AI development, particularly as these models establish themselves as foundation models---large-scale, pre-trained systems that can be adapted for a broad array of downstream tasks \cite{bommasani2021opportunities}. Prominent examples, including GPT-4o, Claude Sonnet 3.5, Gemini 1.5 Pro, and Llama3.1 \cite{ChatGPT, sonnet3.5, Gemini, Llama3.1}, are based on the Transformer architecture \cite{Transformer} and demonstrate high proficiency in retrieval-augmented generation (RAG), computational execution, and data synthesis. Beyond their technical achievements, these Generative AI (GenAI) models are now integral to educational AI applications, transforming learning experiences by enabling personalized support, intelligent tutoring, and resource generation \cite{ahmad2023data, qadir2023engineering}.

\subsection{How LLM-based Educational AI can Embody Biases?}

The foundational nature of LLMs, while beneficial for scalability across tasks, also enables the propagation of inherent biases, impacting various applications and raising particular concerns in educational settings. As these models become integral to shaping learning environments globally, especially through educational platforms like KhanAmigo and Harvard's CS50 AI tutor \cite{KhanAmigo, CS50_AI_TUTOR}, it becomes crucial to evaluate not only their technical performance but also their alignment with ethical and cultural values. Many LLMs are trained on predominantly secular, materialistic, and reductionist datasets and benchmarks \cite{BiasInLLMs}, which can inadvertently lead them to reflect biases that marginalize or overlook diverse cultural, ethical, and spiritual perspectives, especially those outside Western paradigms. These biases, far from being merely theoretical, have been shown to manifest in real-world applications, perpetuating stereotypes or reflecting negative biases toward specific communities \cite{weidinger2021ethical}---an outcome that limits the scope of cultural and ethical diversity presented to young learners.

In global educational contexts, these biases create additional ethical and social challenges. When foundation models are applied across culturally diverse settings without sufficient adaptation, they risk misalignment with local values and worldviews, potentially alienating or misrepresenting certain communities. Recent studies have documented unintended biases in LLMs, including those related to ethnicity, cultural background, and other demographic factors \cite{BiasChats, GhostPaper, BiasPCAPaper}. These findings highlight the limitations of current LLMs in representing the pluralism inherent in human societies, underscoring the need for a framework that actively addresses these biases to support a genuinely inclusive educational AI that respects the diversity of global learners.

\subsection{Using Multiplexity Framework to Analyze LLMs}

In this paper, we propose auditing foundation LLMs through the lens of \textit{Multiplexity}, a framework proposed by Senturk et al. \cite{senturk2020comparative} rooted in Islamic epistemology that encompasses a pluralistic approach to knowledge by incorporating spiritual, ethical, and metaphysical dimensions \cite{AIEthicsPaper}. The multiplexity approach serves as a corrective to the uniplex mindset, which historically emphasizes empirical and secular knowledge, often marginalizing other ways of knowing \cite{qadir2024educating}. As noted by the authors in their work \cite{AIEthicsPaper} and articulated by Senturk \cite{senturk2020comparative}, multiplexity moves beyond a single-dimensional view of knowledge by embracing multiple layers of understanding, each valued for its unique perspective and contribution. Table \ref{table:uniplex_multiplex_comparison} contrasts the uniplex and multiplex worldviews, highlighting key differences in knowledge, ethics, and cultural diversity, with the multiplex approach offering a more inclusive framework for culturally aligned AI \cite{senturk2020comparative, AIEthicsPaper}.

\begin{table*}[!t]
\centering
\caption{Comparison of Uniplex and Multiplex Worldviews, highlighting fundamental differences in approaches to knowledge, ethics, and cultural perspectives. Adapted from \cite{senturk2020comparative, AIEthicsPaper}.}
\begin{tabular}{|p{3.3cm}|p{7cm}|p{7cm}|}
\hline
\textbf{Aspect} & \textbf{Uniplex Worldview} & \textbf{Multiplex Worldview} \\
\hline
\textbf{Foundational Philosophy} & Rooted in Enlightenment thinking, prioritizes empirical and material knowledge. Focuses on objective, secular perspectives. & Embraces diverse epistemological sources, including empirical, spiritual, ethical, and metaphysical dimensions. \\
\hline
\textbf{Knowledge Scope} & Narrow, often reductionist; emphasizes scientific and technical knowledge while sidelining ethical and spiritual dimensions. & Holistic and inclusive; integrates multiple layers of knowledge, balancing technical, ethical, and spiritual insights. \\
\hline
\textbf{Approach to Knowledge} & Views knowledge as value-neutral and often universal, focusing on control and predictability. & Sees knowledge as culturally and ethically situated, emphasizing understanding and wisdom over control. \\
\hline
\textbf{Education Goals} & Primarily aims to develop technical skills and problem-solving abilities for economic productivity. & Seeks to foster holistic human development, encouraging ethical, spiritual, and intellectual growth. \\
\hline
\textbf{Ethical Framework} & Limited or implicit; values tend to align with materialistic and utilitarian ethics, often disregarding virtues or cultural values. & Explicitly includes ethics and values, emphasizing virtues and moral development across cultures. \\
\hline
\textbf{View of Human Potential} & Often mechanistic; sees humans as rational beings primarily defined by cognitive skills. & Recognizes a multi-dimensional view of humans as beings with intellectual, spiritual, and ethical capacities. \\
\hline
\textbf{Cultural Diversity} & Tends to universalize Western norms and downplays or ignores non-Western perspectives. & Embraces pluralism and respects the diversity of cultural and philosophical traditions. \\
\hline
\textbf{Application in AI} & Focuses on developing efficient, powerful AI systems with less emphasis on ethical or cultural context. & Prioritizes the development of culturally sensitive AI that aligns with diverse ethical frameworks. \\
\hline
\textbf{Outcome in Society} & May lead to monocultural or one-size-fits-all solutions, often neglecting local contexts and values. & Promotes inclusivity and adapts solutions to local cultures, fostering a more resilient and diverse society. \\
\hline
\end{tabular}
\label{table:uniplex_multiplex_comparison}
\end{table*}

\begin{figure}[!t]
    \centering
    \includegraphics[width=.8\columnwidth]{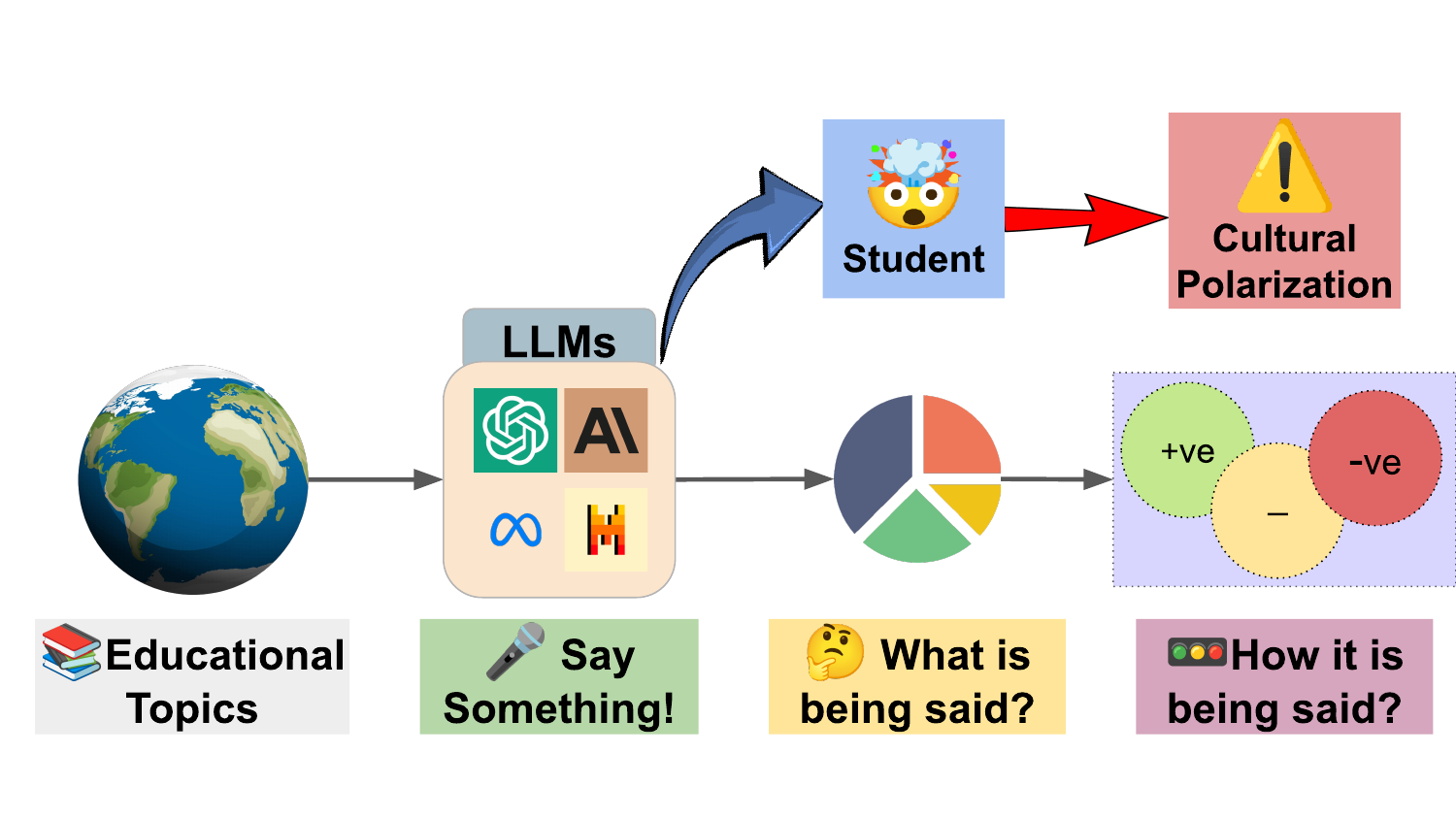}
    \caption{Overview of a multiplexity-inspired framework for assessing cultural bias in LLM-generated educational content: \textit{a two-stage process analyzing cultural distribution across topics and using sentiment analysis to identify biases for a holistic view of cultural alignment.}}
    \label{fig:Framework_Overview}
\end{figure}

This study builds on the multiplexity framework to establish a robust evaluation system for assessing foundation LLMs in terms of cultural and ethical alignment across a wide range of intellectual traditions, including Western, Eastern, Islamic, African, and Latin American perspectives. Our framework systematically scans for indications of monoculture, Eurocentrism, and uniplexity---biases that often prioritize singular, predominantly Western viewpoints---thus addressing potential imbalances in LLM-generated content. By employing a Multi-Agent System (MAS), where agents specializing in distinct cultural viewpoints provide insights, our framework enables a balanced synthesis of perspectives through a central Multiplex Agent that integrates the principles of multiplexity. This approach, illustrated in Figure \ref{fig:Framework_Overview}, enables LLMs to generate content that is pluralistic and inclusive. Such a framework is particularly valuable in educational settings, where ensuring cultural resonance and ethical alignment is essential for fostering inclusive and diverse learning experiences.

\subsection{Contributions of this Paper}

While previous studies have examined biases in LLMs, this paper focuses specifically on educational settings, where ethical, cultural, and local values are essential for holistic learner development. Our contribution is a culturally aligned educational AI framework, evaluating LLMs through a multiplexity-inspired approach. By assessing cultural alignment, we ensure these models respect diverse identities, supporting their suitability for global educational applications. To our knowledge, this is the first study to use a multiplexity framework with LLMs for a pluralistic, education-focused AI approach. This work lays the groundwork for culturally sensitive tools that foster inclusive learning across varied backgrounds.

\subsection{Organization of this Paper}

The rest of the paper is organized as follows. Section \ref{sec:LitReview} reviews relevant literature, covering the development of LLMs, ethical AI frameworks, and existing approaches to cultural bias in AI. Section \ref{sec:Methodology} presents our methodology, detailing the multiplexity-inspired framework for assessing cultural bias in LLM-generated content. Section \ref{sec:Results} discusses the results of applying this framework to various LLMs, and finally, Section \ref{sec:Conclusions} concludes the paper, highlighting key insights and directions for future research.

\section{Literature Review}
\label{sec:LitReview}
The advances in language models through LLMs have significantly reshaped natural language processing (NLP) through unsupervised training on massive corpora. Built initially on simpler models, these architectures have advanced from recurrent neural networks to transformer-based architectures, enabling LLMs to generalize beyond task-specific fine-tuning and perform effectively on diverse linguistic tasks. With this advancement, however, a critical need has emerged to scrutinize biases and ethical considerations.

\subsection{Bias and Ethics in LLMs}
Bias in language models is a significant and well-documented concern, affecting diverse fields such as politics, healthcare, religion, and education. Researchers categorize the associated harms into two primary types: \textit{Allocation Harm}, where an AI system results in an unequal distribution of resources or opportunities, and \textit{Representation Harm}, where a model reinforces stereotypes that impact societal perceptions of certain groups. As analyzed in \cite{BiasChats}, the origins of bias in LLMs can be traced to the pretraining phase, user interactions, and the broader context of the model's deployment. 

A notable example of allocation harm occurs in hiring algorithms that inadvertently favor certain demographic groups over others, leading to unequal job opportunities. For instance, recruitment models trained on biased data may underrepresent qualified candidates from marginalized backgrounds, thereby perpetuating systemic inequalities. Meanwhile, an example of representation harm is documented with OpenAI's GPT-3, which, in analogy-based completions involving Muslims, generated violent associations at a disproportionately high rate (66\%) compared to other religious terms, as noted in \cite{BiasInLLMs}. These examples illustrate that, despite ongoing mitigation efforts, achieving complete neutrality is a persistent challenge, underscoring the need for continuous ethical review and a comprehensive framework to address and mitigate these imbalances in representation and power.

Cultural values shape both individuals and communities, making standardized, one-size-fits-all approaches unsuitable for addressing cultural biases in LLMs. Using the Moral Values Pluralism (MVP) framework, researchers assessed GPT-3's alignment with diverse cultural narratives and found a dominant US- and Eurocentric bias, likely stemming from its predominantly English-language training data \cite{GhostPaper}. As a result, models like GPT-3 often reflect American cultural and moral values, highlighting the challenge of creating AI that fairly represents diverse perspectives.

To address cultural biases, \cite{BiasPCAPaper} proposed persona-based prompts, encouraging LLMs to adopt perspectives aligned with specific cultural identities. Results suggested variances in moral and cultural expressions across different LLMs, with GPT-4 displaying a more secular worldview than GPT-4-turbo, which tended towards traditional values. These studies show how there's a limitation of these LLMs in representing global cultural values, thus requiring the addressing of such concerns. 

\subsection{LLMs in Education}
\label{subsec:LLMs_in_Education}

The integration of AI in education, particularly through LLMs \cite{KhanAmigo, CS50_AI_TUTOR}, has led to frameworks emphasizing ethical alignment and cultural sensitivity. One such framework, Multiplex AI Humanities (MAIH) \cite{qadir2024educating}, critiques the uniplex worldview---rooted in reductionist, secular, and empirically-driven Enlightenment ideals---that often universalizes Eurocentric perspectives at the expense of global diversity.

In contrast, MAIH offers a pluralistic, open-civilizational framework that values both empirical and metaphysical dimensions, incorporating diverse cultural, intellectual, and religious perspectives. It is built on two pillars \cite{senturk2020comparative}: \textit{Multiplex Ontology}, which recognizes physical, metaphysical, and divine dimensions; and \textit{Multiplex Epistemology}, which includes diverse forms of knowledge, such as reason, intuition, and revelation. By respecting the multi-layered connections between technology, knowledge, and the human condition, multiplexity fosters inclusive and adaptable education that aligns with the diverse values and aspirations of communities worldwide.

Furthermore, practical applications of LLMs in education benefit from ``practical wisdom'' (\textit{phronesis}), as proposed in \cite{PhronesisPaper}, to promote not only technical proficiency but also ethical reasoning and cultural awareness. Sawalha et al. \cite{sawalha2024analyzing} investigate how students employ ChatGPT for problem-solving, identifying effective prompting strategies that enhance accuracy and learning outcomes, with observed differences based on gender. Similarly, Johri et al. \cite{johri2023generative} discuss the transformative impact of generative AI tools in engineering education, emphasizing the importance of `sense and knowledge checking' to prevent reliance on AI without critical oversight. Johri \cite{adityaJohriQNA} further suggests that AI should not be viewed as a monolithic entity but as a flexible tool that requires contextual understanding to meet the needs of diverse communities.

Building on the multiplexity framework, our work presents a structured evaluation system to identify and mitigate cultural biases within LLM-generated content. By examining the educational outputs of LLMs for signs of monoculturalism, Eurocentrism, and uniplexity, we aim to support a culturally inclusive approach in AI education. Using a MAS where agents specializing in distinct cultural viewpoints provide insights to a central Multiplex Agent, our framework integrates diverse perspectives to produce pluralistic, ethically aligned educational content. This approach is essential for fostering inclusive learning experiences that respect the cultural diversity of educational settings. In the Methodology section (\S \ref{sec:Methodology}), we further discuss how we applied and implemented multiplexity principles to make LLMs more culturally inclusive.

\begin{figure*}[t]
    \centering
    \includegraphics[width=0.85\textwidth]{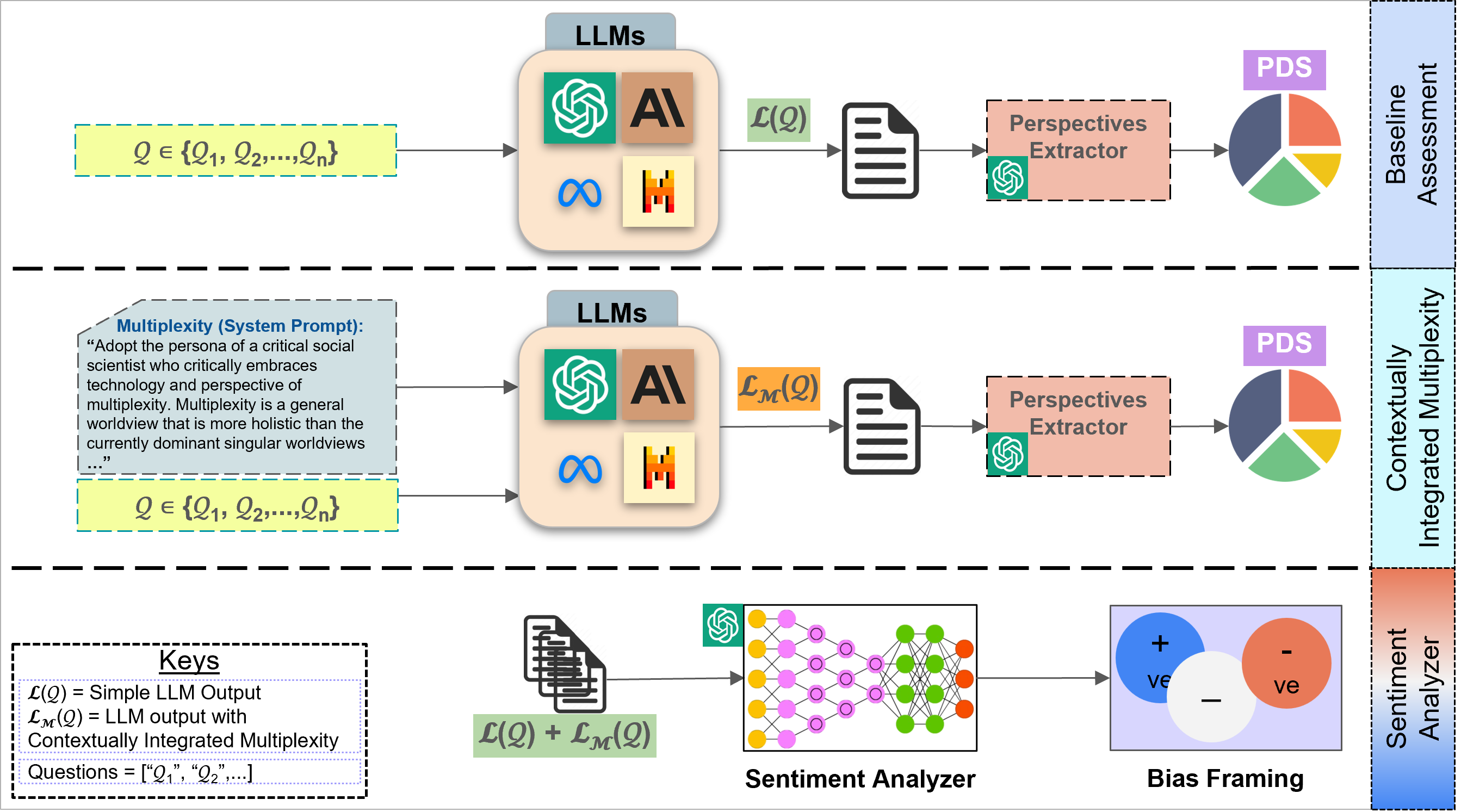}
    \caption{\textbf{System design for utilizing contextually-implemented multiplex LLMs to enrich baseline LLM}. \textit{\textbf{Baseline Assessment:} LLMs respond to educational questions, and their responses are analyzed by the Perspective Extractor to identify cultural references. The Perspective Distribution Score (PDS) then quantifies the cultural representation within these responses. 
    \textbf{Contextually-implemented Multiplexity}: LLMs answer similar questions with Multiplexity prompts, designed to incorporate diverse cultural perspectives. The Perspective Extractor analyzes these responses, and PDS scores are calculated to assess cultural integration.
    \textbf{Sentiment Analysis:} The sentiment and tone of LLM responses for each culture are assessed through zero-shot classification by GPT-4o, enabling bias framing across cultural perspectives.}}
    \label{fig:Pipeline1_2}
\end{figure*}


\section{Methodology}
\label{sec:Methodology}

In this study, we introduce a novel multiplexity-based AI evaluation framework designed to assess the cultural alignment of LLMs across perspectives such as Western, Eastern, Islamic, Latin American, etc. Concretely, the framework comprises three distinct strategies (1) baseline assessment; and two kinds of multiplex strategies: (2) Contextually-Implemented Multiplexity, and (3) Multi-Agent System (MAS)-Implemented Multiplexity, each employing the Perspectives Extractor, Perspective Distribution Score (PDS), and Sentiment Analyzer/ Bias Framing (see \S \ref{sec:System_Design} for more details) to evaluate the coverage of cultural perspectives in the models' outputs. We developed eight distinct categories of educational topics, each containing a unique set of questions (see Table [\ref{tab:question_distribution}]). In each strategy, LLMs are asked to respond to these questions, and their responses are used to extract references to each cultural perspective, which are then passed to the PDS for final score calculation, and the same responses are passed to the sentiment analyzer for bias framing towards each culture.

\subsection{Research Questions} \label{subsec:RQs}
This study employs ``applied multiplexity'' to create an evaluation system that assesses foundation LLMs for cultural and ethical alignment across diverse intellectual traditions, including Western, Eastern, Islamic, African, and Latin American perspectives. By implementing a MAS with agents specializing in distinct cultural perspectives, the framework synthesizes insights through a central Multiplex Agent, harmonizing outputs into balanced, inclusive responses. Initially, baseline cultural bias is evaluated via perspective coverage (Fig. \ref{fig:Pipeline1_2}), followed by applying multiplexity-informed contextual rules to ensure balanced representation. This approach enables LLMs to generate culturally resonant and ethically aligned content, which is particularly valuable in educational contexts where inclusive learning is essential. The resulting outputs support holistic educational experiences, addressing the diverse needs of learners and educators globally.

Specifically, we address three research questions (RQs):

\vspace{2pt}

\textbf{RQ1:} \emph{What forms and levels of cultural bias are present in LLMs by default, and how can these biases be effectively measured as a baseline?}

\textbf{RQ2:} \emph{How can the Multiplexity Framework, along with a Multiplexity-inspired Multi-Agent System, be utilized to create a pluralistic approach in educational AI?}

\textbf{RQ3:} \emph{How can the Multiplexity Framework be applied to guide LLMs in fostering positive and respectful representations of diverse cultural perspectives?}

\vspace{2pt}

\subsection{Cultural Analysis and Sentiment Assessment Strategies}

\vspace{2mm}\subsubsection{Strategy 1: Baseline Assessment}

The first strategy assesses the baseline performance of various LLMs concerning cultural bias. This strategy evaluates the coverage of different cultures in the responses of these LLMs out of the box. The PDS is utilized to measure how well each perspective is represented in the models' responses, allowing for a comprehensive understanding of the cultural biases present. No additional information is provided to the models except for the questions. The output of the model is also used to extract the sentiment of each LLM towards these perspectives for bias framing (see \S \ref{subsec:strategy1} and Fig. \ref{fig:Pipeline1_2} for more details).

\vspace{2mm}\subsubsection{Strategy 2: Contextually Implemented Multiplexity}
The second strategy integrates the ideas of multiplexity to provide contextual guidelines that promote balanced outputs across cultures. The core concepts of multiplexity and its rules are provided in the system prompt (context) of the models to follow for their responses. These responses are passed to the PDS to evaluate the diversity and representation of perspectives, ensuring that the outputs generated adhere to the principles of cultural inclusivity along with the sentiment of the model for bias framing and comparison with baseline assessment (More details in \S \ref{subsec:strategy2} and Fig. \ref{fig:Pipeline1_2}).

\vspace{2mm}\subsubsection{Strategy 3: MAS-Implemented Multiplexity}

The final strategy involves implementing a multi-agent pipeline, where agents specializing in distinct cultural perspectives give responses that are then synthesized by a Multiplex Agent according to the multiplexity principles. Each agent is asked to answer the questions with respect to its culture and these responses from agents are passed to a multiplex agent to apply the multiplexity principles ensuring a holistic viewpoint of producing knowledge. The PDS is applied to assess the coverage of each cultural perspective on the response of multiplex agent, helping to ensure a harmonious integration of diverse viewpoints. Again, the output of this system is also utilized for bias framing by analyzing the sentiment of the system towards each perspective. (see \S \ref{subsec:strategy3} and Fig. \ref{fig:Pipeline3} for more detailed implementation and working.) 

In all these strategies, the \textit{PDS and \textit{Sentiment Analyzer/Bias Framing}} are central components that enable the framework to effectively evaluate the cultural alignment of LLMs and promote more holistic educational content. We discuss the possibility of applying these approaches during the post-training phase of the LLMs to ensure that the models are culturally inclusive and respectful of diverse viewpoints, particularly in an educational setting.

\subsection{System Design \& Implementation Details}
\label{sec:System_Design}
In this section, we will discuss the system design of our proposed framework for all three strategies mentioned in the methodology section. We designed a self-contained system for this evaluation purpose, meaning that no external datasets or information sources are used except for the LLMs themselves. This system design ensures that there are no external factors contributing to the cultural assessment, focusing solely on the LLMs. Figure \ref{fig:Pipeline1_2} shows the overall structure of Strategy One---\textit{Baseline Assessment}---and Strategy Two---\textit{Contextually Implemented Multiplexity} while Figure \ref{fig:Pipeline3} illustrates the strategy of using \textit{MAS-Implemented Multiplexity}. 

For this framework, we selected eight distinct cultural perspectives representing various regions of the world. This selection aims to ensure that the scores in our results are comprehensive and well-rounded. The chosen cultural perspectives are \textbf{(1)} Western \textbf{(2)} Eastern \textbf{(3)} Islamic \textbf{(4)} African \textbf{(5)} Latin American \textbf{(6)} Indigenous \textbf{(7)} South Asian and \textbf{(8)} Others. The ``Others'' category allows us to incorporate all remaining cultures into one classification to make this study feasible. 

We used both open-source and closed-source/proprietary models which include \textbf{(1)} GPT-4o \textbf{(2)} Claude 3.5 Sonnet and \textbf{(3)} Llama 3.1 8B \textbf{(4)} Mistral 7B. All of the models are used with the latest checkpoints. As discussed earlier, we created eight different categories of topics $\mathcal{T}$, with each category containing different questions $\mathcal{Q}$ on the same subject. Table \ref{tab:question_distribution} presents the selected categories of topics and the number of questions per topic category used in our system. The questions are designed to be diverse both within each category and across all categories to cover a broad range of educational content with the help of professional educators from different universities around the world.

\begin{table}[htbp]
    \centering
    \caption{Categories and Number of Questions}
    \resizebox{\columnwidth}{!}{ 
        \begin{tabular}{|l|c|} 
            \hline
            \rowcolor[HTML]{DAF7A6}
            \textbf{Topics Category $\mathcal{T}$} & \textbf{Number of Questions $\mathcal{Q}$} \\ 
            \hline
            Mathematical Topics                        & 5 Questions     \\ 
            \hline
            Economical Topics                          & 5 Questions     \\ 
            \hline
            Design Topics                              & 5 Questions     \\ 
            \hline
            Lab Related Topics                         & 5 Questions     \\ 
            \hline
            Optimization Topics                       & 5 Questions     \\ 
            \hline
            Social/Political Topics                    & 5 Questions     \\ 
            \hline
            Ethical Topics                             & 5 Questions     \\ 
            \hline
            Philosophical/Historical/Knowledge-based Topics & 12 Questions \\ 
            \hline \hline
            \rowcolor[HTML]{F4CCCC}
            \textbf{Total}                            & \textbf{47 Questions} \\ 
            \hline
        \end{tabular}
    }
    \label{tab:question_distribution}
\end{table}

These questions are used to generate prompts which are then passed to the LLMs for their responses. The prompts are designed in such a way that the LLMs only provide the desired output and no extra information, commentary, or preambles. Once passed to the LLMs, they generate their responses and these responses are further processed for PDS and sentiment analysis. Box \ref{pabox:Prompt_template} shows the template used for generating prompts for the LLMs. These prompts are used for both Baseline Assessment (\S \ref{subsec:strategy1}) and Contextually Implemented Multiplexity (\S \ref{subsec:strategy2}) strategies. The details regarding the MAS are shared in \S \ref{subsec:strategy3}.

\begin{figure}[!t]
    
\begin{pabox}[label=pabox:Prompt_template]{Prompt Template for Models}
\footnotesize
\ttfamily 
""" Write \textbf{\{MAX\_WORDS\}} words on the following question: "\textbf\{Question\}". Just give me your output and nothing extra. Do not produce any garbage or random text just accurate answers. Your output is: """
\end{pabox}
\end{figure}

\vspace{2mm}\subsubsection{Strategy 1: Baseline Performance Assessment} 
\label{subsec:strategy1}
In this assessment strategy, we established a baseline to compare with the proposed mitigation strategies Fig. \ref{fig:Pipeline1_2} (first row of figure). LLMs are used in their original form to ensure that no bias or external information is influencing the raw responses. Prompts are generated according to the template in Box \ref{pabox:Prompt_template} and are provided to the LLMs to elicit responses $\mathcal{L(Q)}$. These responses are then passed to the Perspectives Extractor to extract references made to different cultures (see \S\ref{subsec:Perspectives_Extractor}), the PDS to calculate the distribution score based on references (see \S\ref{subsec:Perspectives_Distribution_Extractor}), and the Sentiment Analyzer (see \S\ref{subsec:Sentiment_Analyzer}) to extract the sentiment of these LLMs towards each culture.

\vspace{2mm}\subsubsection{Strategy 2: Contextually Implemented Multiplexity} 
\label{subsec:strategy2}
In this strategy, we crafted a system prompt that embodies the principles of Multiplexity, guiding the LLMs to operate with a nuanced, pluralistic perspective. The prompt instructs the model to adopt the persona of a critical social scientist who, while open to technological advancements, remains aware of their potential societal impacts and limitations. The second row of Figure \ref{fig:Pipeline1_2} shows the proposed methodology for this strategy.

This approach incorporates key concepts from Multiplexity, including \emph{Multiplex Ontology}, \emph{Multiplex Epistemology}, as detailed in \S \ref{subsec:LLMs_in_Education} and proposed by authors in \cite{senturk2020comparative}. Our contextual design aims to foster a decision-making process that integrates diverse perspectives, respects cultural and philosophical multiplicity, and emphasizes human dignity and ethical conduct. By promoting ``both-and'' thinking, this strategy encourages the model to engage in complex analyses that consider historical contexts, the wisdom of diverse traditions, and a universal outlook that values human fraternity and rights.

LLMs with this system prompt are used with the same questions $\mathcal{Q}$ across all topic categories $\mathcal{T}$, employing the same prompts as in the baseline strategy with the same prompt template \ref{pabox:Prompt_template} to generate responses $\mathcal{L_M(Q)}$. These responses are then processed by the Perspectives Extractor, the PDS, and the Sentiment Analyzer for bias framing.

\vspace{2mm}\subsubsection{Strategy 3: MAS-Implemented Multiplexity} 
\label{subsec:strategy3}
Recent studies \cite{huang2023agentcoder, hong2023metagpt, ghafarollahi2024protagents, yu2023co, tang2023medagents} show that LLMs achieve superior performance on complex tasks when utilizing a collection of agents collaborating on different aspects of a problem, compared to using a single LLM. The concepts of collective intelligence and a collaborative approach to problem-solving, with agents negotiating to achieve an optimized solution, have demonstrated promising results. In this study, we proposed a strategy that employs Camel AI \cite{li2023camel} framework to create an LLM-based MAS that is multi-cultural by following Multiplexity. Camel AI utilizes inception prompting techniques as part of its framework designed to facilitate task completion by guiding chat agents while ensuring that their responses remain consistent with human intentions.

\begin{figure*}[htbp]
    \centering
    \includegraphics[width=0.85\linewidth]{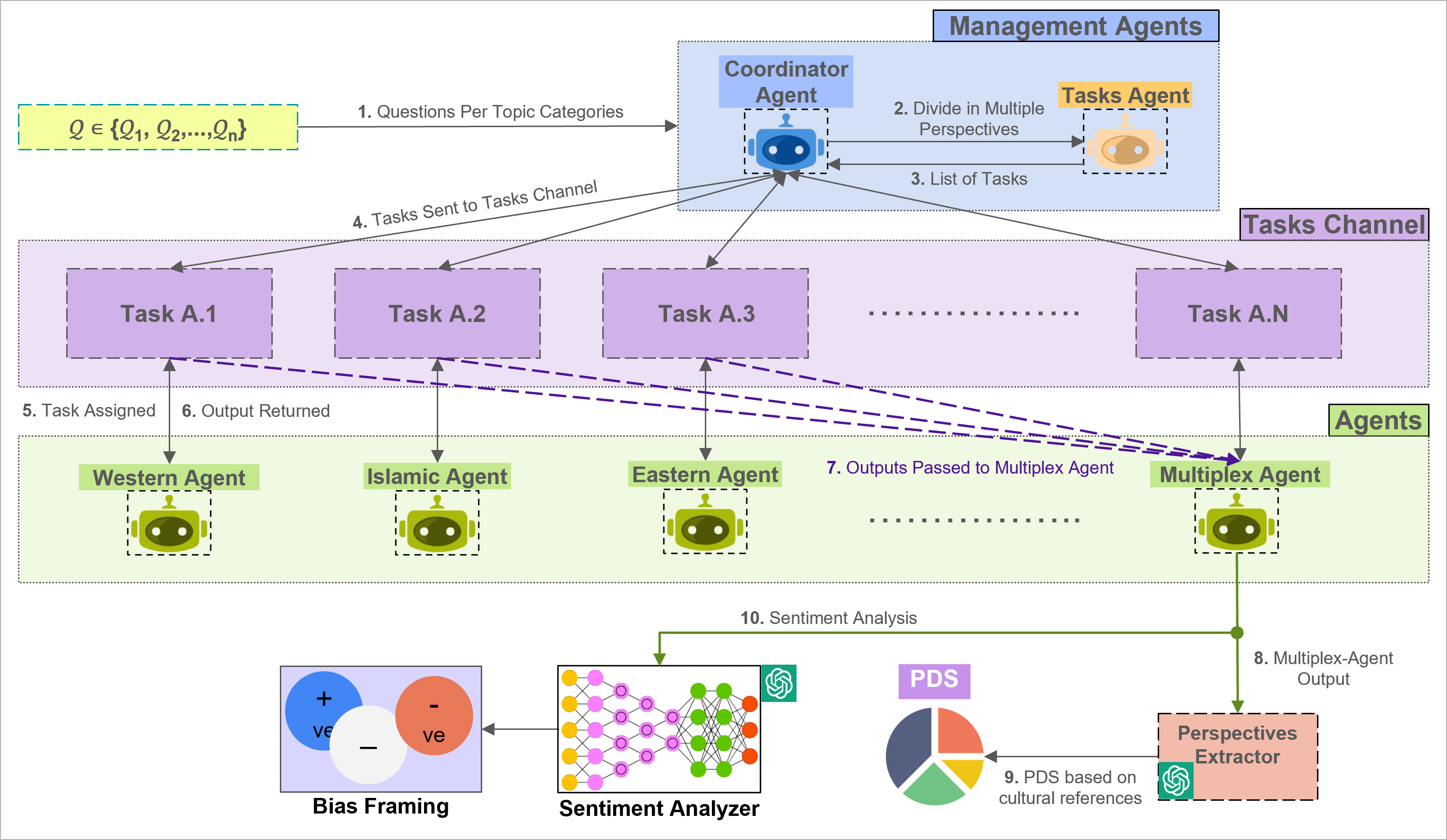}
    \caption{\textbf{System design for utilizing a Multi-Agent System to enrich LLM responses with multicultural perspectives}. 
    \textit{\textbf{1)} Questions on educational topics are sent to the Coordinator Agent. 
    \textbf{2)} The Coordinator forwards them to the Tasks Agent to generate a task list. 
    \textbf{3)} The Tasks Agent returns the task list to the Tasks Channel. 
    \textbf{4)} Tasks are sent to the Tasks Channel.
    \textbf{5)} Tasks get assigned to relevant agents (excluding the Multiplex Agent) based on their personas.
    \textbf{6)} Agents generate outputs, sending them to the Tasks Channel and Coordinator.
    \textbf{7)} The Coordinator sends all outputs to the Multiplex Agent, which applies Multiplexity rules for a multicultural output.
    \textbf{8)} The Perspectives Extractor identifies cultural references in this output.
    \textbf{9)} The PDS is calculated from these references.
    \textbf{10)} A Sentiment Analyzer assesses sentiment toward each culture and results are reviewed to evaluate bias framing.}
    }
    \label{fig:Pipeline3}
\end{figure*}

Our system design, illustrated in Figure \ref{fig:Pipeline3}, leverages Camel AI's WorkForce to create a structured society of agents that collaboratively address complex tasks across diverse cultural contexts. The WorkForce employs a coordinator agent to facilitate task decomposition and distribution based on each agent's expertise. This coordinator agent, in conjunction with a task agent, dynamically assesses agent performance, identifying cases where outputs do not meet predefined standards or where repeated task failures occur. In such instances, the system adapts by reassigning tasks to agents with appropriate expertise or spawning new agents with specialized cultural focus, enhancing both task efficiency and system reliability. Each agent is developed with a culturally aligned persona (Box \ref{box:Agent_Persona}), carefully crafted to reflect specific cultural perspectives relevant to our broad question set $\mathcal{Q}$ across categories of educational topics $\mathcal{T}$. Rather than creating static prompts (Box \ref{pabox:Prompt_template}), we utilize Camel AI's Task functionality to dynamically generate culturally relevant tasks for WorkForce agents (Box \ref{box:WorkForce_Task}). Each agent provides responses that reflect its assigned cultural persona and task requirements, which are then synthesized by the Multiplex Agent, guided by the Multiplexity system persona into a unified, multicultural output aimed at enriching educational content. This consolidated output, $\mathcal{L_{MA}(Q)}$, is subsequently evaluated by the Perspectives Extractor, PDS, and Sentiment Analyzer to assess the breadth of cultural perspectives and identify any framing biases.

\begin{figure}[!t]
    \begin{pabox}[label=box:Agent_Persona]{Persona of Islamic Agent used in MAS}
    \footnotesize
    \ttfamily 
    \textbf{"Islamic Agent"}: """ You are an AI assistant representing Islamic values centered on faith, morality, and justice derived from Islamic teachings. In historical, philosophical, or ethical discussions, you reference the Quran, Hadith, and scholars like Al-Ghazali. For topics like mathematics, design, or economics, your focus shifts to relevant principles, practices, and techniques, ensuring that responses remain practical and context-appropriate, avoiding direct cultural references but you can make cultural references as long as they are relevant to the technical questions. """
    \end{pabox}
\end{figure}

\vspace{2mm}\subsubsection{Perspectives Extractor} \label{subsec:Perspectives_Extractor}
The Perspectives Extractor is based on GPT-4o \cite{ChatGPT}, which is used to extract all references made in the responses of the LLMs. The LLMs' responses are passed to this model to detect any references to different cultural perspectives within the output, both at the surface level and contextually based on the scenario. It is important to use a classification model that can understand content both contextually and the intent behind the content to analyze and classify the parts of the content which are referred to a culture. Since no pre-trained models are available for this specific task, we utilized GPT-4o with its impressive performance on zer-shot classification for this purpose.

\begin{figure}
    
\begin{pabox}[label=box:WorkForce_Task]{Multi-Agent LLMs WorkForce Task Template}
\footnotesize
\ttfamily 
""" You are given different agents representing various cultural perspectives. You are tasked to write exactly \textbf{\{MAX\_WORDS\}} words on the topic \textbf{\{"TOPIC"\}}. \\
First, ask each agent to write \textbf{\{MAX\_WORD\_PER\_AGENT\}} words of content on the topic except for the Multiplex agent. Then, take the output of all the agents and pass it to the Multiplex Agent to write a synthesis of the different perspectives presented by all the other agents based on the rules of multiplexity that are already given to this agent. """
\end{pabox}

\end{figure}

For this step, we designed a prompt template that takes the LLMs' responses and the selected cultural perspectives, asking the model to return a Python dictionary containing all references to each cultural perspective. We have found that asking GPT-4o to provide a Python dictionary for references is more consistent and easier to manage than a text-based output, which can be inconsistent and difficult to parse.

\vspace{2mm}\subsubsection{Perspectives Distribution Extractor (PDS)} \label{subsec:Perspectives_Distribution_Extractor}
The output of the Perspectives Extractor (\S \ref{subsec:Perspectives_Extractor}) is passed onto this part of the framework which parses the Python dictionary and assigns scores to each cultural perspective. 

\paragraph{PDS} The PDS is a metric that quantifies the proportional prominence of each cultural perspective within a set, measuring each perspective's share of the total reference count across all perspectives. The PDS is represented as a vector \(\mathbf{PDS} = [P_1, P_2, \ldots, P_n]\), where each element \(P_i\) corresponds to the score of a specific perspective \(P_i\). The score for each perspective is defined as:

\[
P_i = \frac{R_i}{\sum_{j} R_j}
\]

where \(R_i\) represents the reference count for perspective \(P_i\), and \(\sum_{j} R_j\) is the total reference count for all perspectives in the dictionary. By design, the PDS for all perspectives sum to 1. This provides a normalized measure that allows for a straightforward comparison of each perspective's prominence as a fraction of the total distribution. A score closer to 1 indicates a highly prominent perspective, while scores near 0 reflect perspectives with lesser visibility.

\paragraph{PDS Entropy} The PDS Entropy score measures how evenly cultural perspectives are represented, with high entropy indicating balanced diversity and low entropy indicating less diversity or dominance by a few cultures. The entropy score \( S \) is calculated as follows:

\vspace{-3mm}
\[
H = - \sum_{i=1}^{n} p_i \log(p_i)
\]
where \( p_i \) represents the individual scores in the \( \texttt{PDS} \) vector and \( n \) is the total number of scores. The normalized entropy score is:

\[
S = \frac{H}{\log(n)} \quad \text{if} \quad \log(n) > 0, \quad \text{else} \quad S = 0
\]

\vspace{2mm}\subsubsection{Sentiment Analyzer - Bias Framing} \label{subsec:Sentiment_Analyzer}
The Sentiment Analyzer is a GPT-4o-based zero-shot classifier. Sentiment analysis assesses the sentiment or style of each LLM toward different cultures. The LLMs' responses are passed to GPT-4o for sentiment classification. For the same reasons as with Perspectives Extraction, we can only use GPT-4o for this purpose. We designed a prompt template for this part of the pipeline. Again, for consistency, we asked the model to return a Python dictionary. 

Sentiment analysis helps us understand how LLMs treat each culture. Generally, LLMs are unlikely to exhibit negative sentiments, as they are post-trained to minimize hate speech or targeted language. While responses may vary depending on the nature of the question, negative references to any entity are rare in the model's output due to these safeguards. However, a neutral sentiment does not necessarily indicate a positive stance toward a particular culture; it may also imply that the LLM does not consider that culture important or worthy of positive regard. Thus, if an LLM's response is neutral toward a culture, it likely means it is not treating that culture equally.


\begin{figure}[h!]
    \centering
    \begin{subfigure}[t]{0.48\textwidth}
        \centering
        \includegraphics[width=\linewidth]{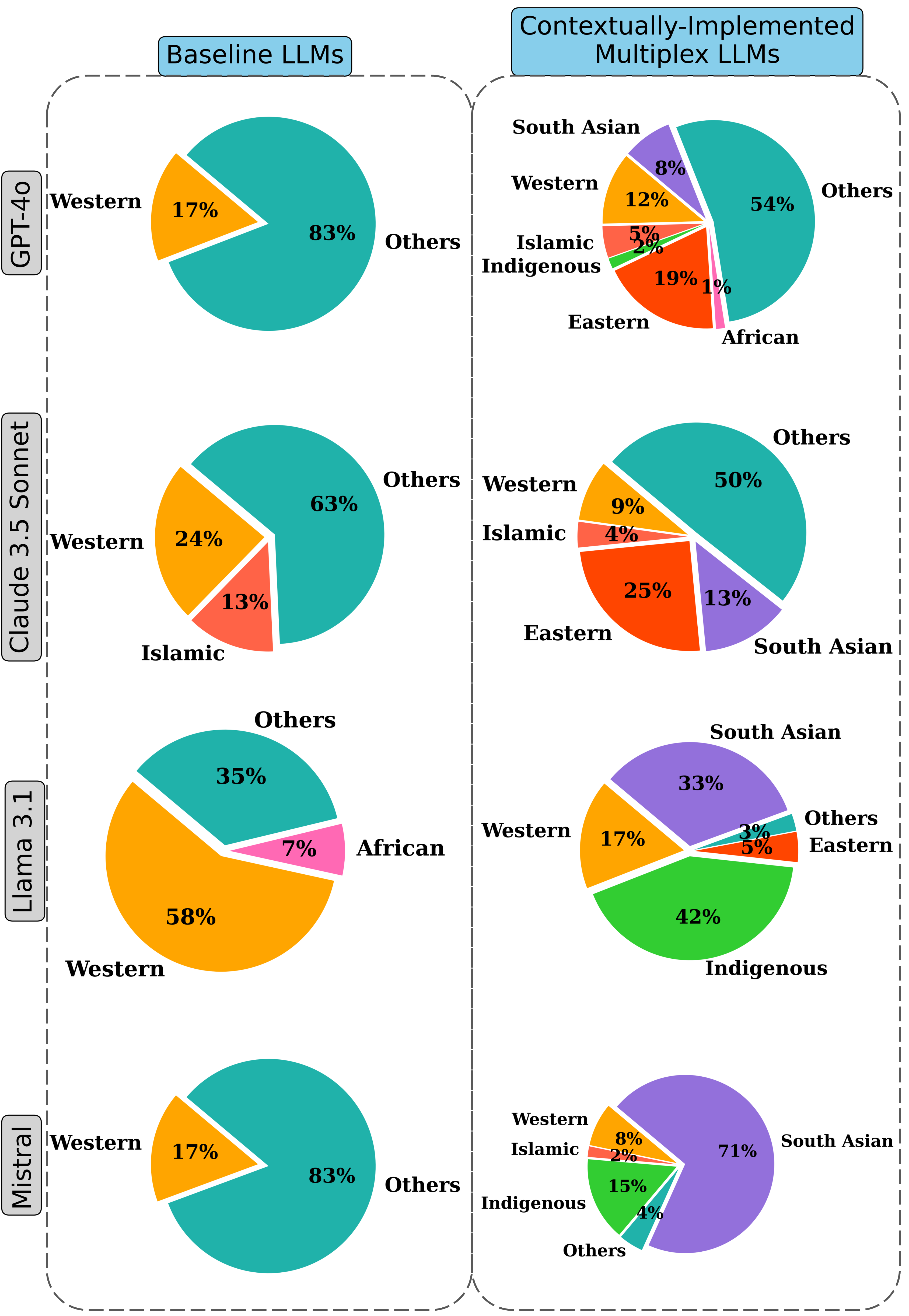}
        \caption{\textit{Baseline LLMs (left column) and the contextually-implemented multiplex model (right column).}}
        \label{fig:results_combined}
    \end{subfigure}
    \vspace{5mm}
    
    \begin{subfigure}[t]{0.33\textwidth}
        \centering
        \includegraphics[width=\linewidth]{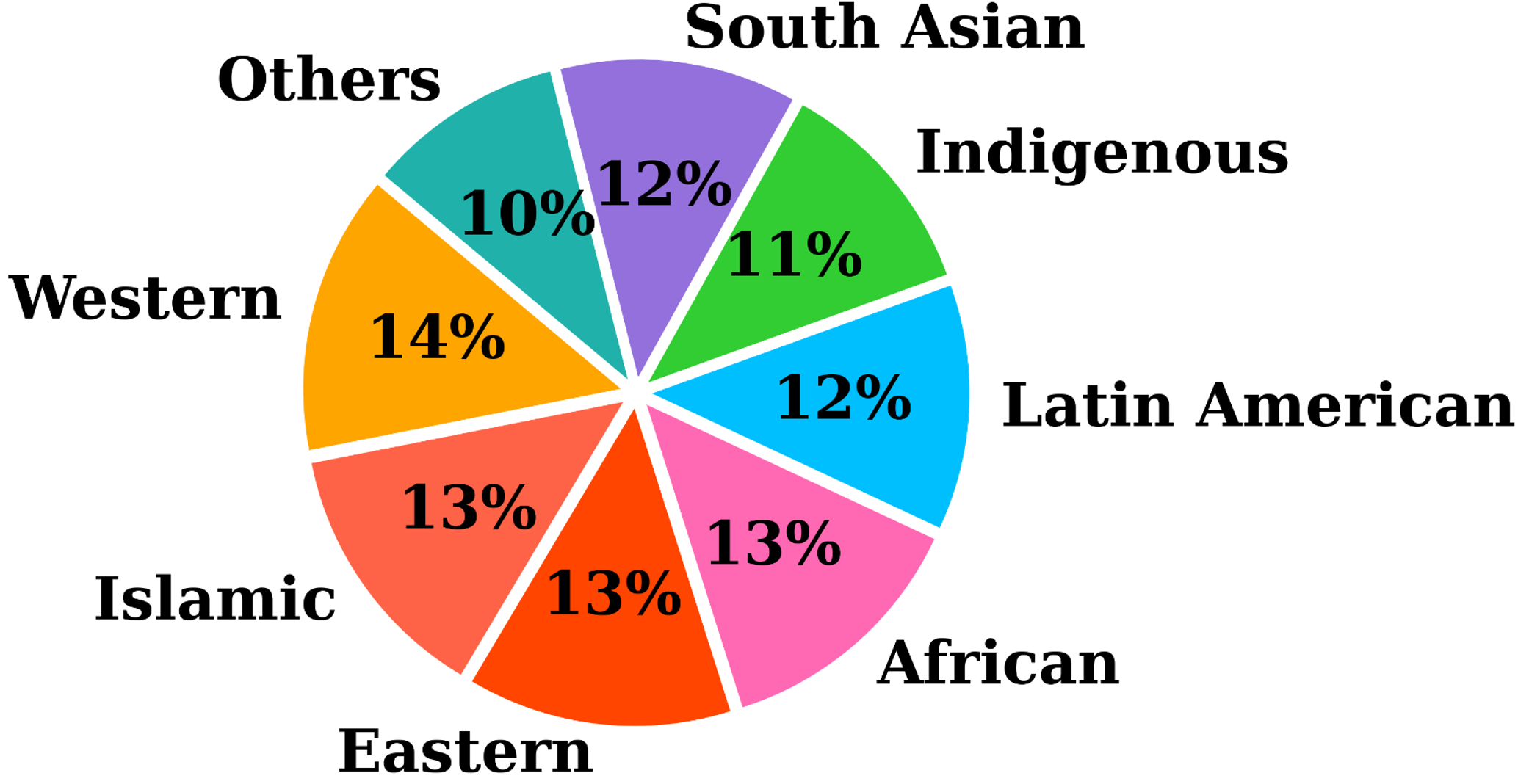}
        \caption{\textit{PDS for MAS-implemented multiplex model (with GPT-4o) performs better than contextually-implemented multiplex model.}}
        \label{fig:MAS_results}
    \end{subfigure}
    \caption{PDS for the baseline LLMs, and the two multiplex adaptations of these LLMs: the contextually-implemented multiplex models and the MAS-implemented multiplex models. \textit{It is clear that multiplex models are more inclusive with MAS-implemented multiplex models performing better. A lack of diversity in the baseline models is also visible. }}
    \label{fig:combined_PDS}
\end{figure}

\section{Results}
\label{sec:Results}

In this section, we examine and discuss the results generated using our methodology to address the three research questions introduced in \S\ref{subsec:RQs}. The findings of our study are summarized in Fig. \ref{fig:combined_PDS} (a visual overview of PDS and multi-cultural coverage) and Fig. \ref{fig:SentimentAnalysis_combined} (breakdown of sentiment scores of uniplex and multiplex models). These results will show that the multiplex models are effective in being inclusive and diverse. We will discuss our results with reference to each of our RQs next.

\subsection{RQ1: Baseline Performance and Cultural Bias Assessment}
To evaluate the baseline performance of these models in terms of cultural inclusivity, we developed a methodology aimed at uncovering hidden biases in LLM responses. Fig. \ref{fig:Pipeline1_2} shows our proposed strategy for assessing the baseline performance of LLMs, specifically designed to address RQ1. Following the complete pipeline outlined in Section \ref{sec:System_Design}, we compiled the results into visual representations, shown as pie charts, to facilitate understanding. Column 1 (Baseline LLMs) of Fig. \ref{fig:results_combined} and Table \ref{tab:PDS_Entropy} presents these baseline PDS for each of the four LLMs.

Across all graphs in Column 1, a distinct pattern emerges where one or at most two cultures dominate the responses, with Western cultural references overwhelmingly prevalent. This trend is also evident in Table \ref{tab:PDS_Entropy}, which shows the PDS Entropy of just 3.25\% for baseline LLMs. This reveals a narrow cultural perspective in the default LLM outputs, highlighting a Western-centric bias. Such results underscore significant cultural biases in the out-of-the-box LLMs, which are currently applied in various educational tools designed to enhance learning through Generative AI. However, our findings indicate that these LLMs, in their default configuration, heavily favor Western philosophies and ideologies, making them culturally unsuitable for educational applications without further alignment. This bias is particularly concerning for achieving globally pluralistic AI in education, as it suggests a lack of cultural pluralism in the models' knowledge base and reasoning processes.

\begin{table}[!t]
    \centering
    \caption{PDS Entropy of LLMs across each strategy ($\uparrow$ indicates higher is better).}
    \resizebox{\columnwidth}{!}{%
        \begin{tabular}{|c|c|c|c|}
            \hline
            \rowcolor[HTML]{DAF7A6}
            
            \textbf{LLMs} & \textbf{Baseline (\( S \))$\uparrow$} & \makecell{\textbf{Contextually Implemented} \\ \textbf{Multiplexity} (\( S \))$\uparrow$} & \makecell{\textbf{MAS-implemented} \\ \textbf{Multiplexity} (\( S \))$\uparrow$} \\
            \hline
            GPT-4o              & 2\% & 28\% & 98\% \\
            Claude 3.5 Sonnet   & 5\% & 24\% & - \\
            Llama 3.1           & 5\% & 13\% & - \\
            Mistral             & 1\% & 11\% & - \\
            \hline \hline
            \rowcolor[HTML]{F4CCCC}
            \textbf{Average }   & \textbf{3.25\%} & \textbf{19\%} & \textbf{98\%} \\
            \hline
        \end{tabular}%
    }
    
    \label{tab:PDS_Entropy}
\end{table}

\subsection{RQ2: Contextual Prompting \& MAS for Cultural Inclusivity}
\subsubsection{\textit{Contextually Implemented Multiplexity}}

To align the LLMs with diverse cultural viewpoints in educational responses, we embedded the core principles and operational multiplexity principles into the system prompt (context) of each LLM. This approach aimed to mitigate cultural biases by contextualizing the LLMs' responses within a pluralistic framework. Fig. \ref{fig:Pipeline1_2} shows our strategy for implementing Multiplexity to reduce cultural biases. Contextually Implemented Multiplexity (the right column in Fig. \ref{fig:results_combined}) shows the improved performance of multiplex LLMs when responding to educational questions. The pie charts in this column reveal a noticeable increase in the inclusion of diverse cultural perspectives, as evidenced by PDS scores for each culture rising above 0\% and an increase in PDS Entropy in Table \ref{tab:PDS_Entropy} from just 3\% in baseline to 19\% with this approach. This shift demonstrates that multiplex LLMs can dynamically adjust their responses to become more inclusive, integrating references and information from various cultural backgrounds. The impact of multiplexity is further evidenced by the redistribution of cultural perspectives in each graph. We observe that LLMs reduce the overemphasis on certain cultures while expanding their representation of others, indicating a genuine alignment towards balanced, multicultural responses. Multiplexity, at its core, is structured to encourage AI systems to recognize a broader range of knowledge acquisition methods---including reason, intuition, and revelation---alongside empirical evidence, accommodating diverse worldviews. These aspects resonate with many cultural traditions, in contrast to cultures where uniplex, or single-dimensional, thinking may be more common. Through multiplexity, our approach moves toward a globally pluralistic AI framework, making LLMs more suitable for educational purposes by fostering cultural respect and inclusivity in their responses. This framework not only enhances the truthfulness and relevance of AI in multicultural contexts but also supports the goal of developing an inclusive, globally accessible AI system that can serve a variety of educational needs.

\subsubsection{\textit{MAS-Implemented Multiplexity}}
MAS is designed to tackle complex reasoning and problem-solving tasks \cite{huang2023agentcoder, hong2023metagpt, ghafarollahi2024protagents, yu2023co, tang2023medagents}. While answering educational questions may seem like a simple task, the baseline assessment results in Fig. \ref{fig:results_combined} clearly show that achieving cultural inclusion and integrating Multiplexity to improve LLMs for educational contexts is, in fact, a challenging goal. While adding Multiplexity in the system prompt of LLMs leads to the emergence of some new cultural references in the responses, there remains an uneven distribution, with certain cultures still disproportionately represented. This calls for an alternative approach that can reliably address this complex issue of cultural inclusion.

To tackle this, we designed a MAS incorporating multiple agents, each focused on a distinct cultural perspective. A multiplex agent then synthesizes the outputs of these individual cultural agents to create a single, culturally inclusive response (see \S\ref{subsec:strategy3} for a detailed implementation). Fig. \ref{fig:Pipeline3} illustrates the design of this MAS. We implemented this approach using the Camel AI \cite{li2023camel} framework. However, due to some design limitations, Camel AI fully supports only OpenAI's models for all functionalities. Thus, to generate results and assess the performance of this strategy, we used GPT-4o as the backend for all agents. Fig. \ref{fig:MAS_results} and Table \ref{tab:PDS_Entropy} present the performance of this strategy in producing culturally inclusive outputs from LLMs.

The graphs in the figure and table show that this proposed MAS strategy is highly effective (higher PDS) in including diverse cultural perspectives in the final output. In this study, we considered a total of eight cultural perspectives. Ideally, for a precisely multicultural output, the PDS would be evenly distributed at 12.5\% for each culture in the responses. The results indicate that our MAS achieves this target, maintaining a distribution close to $12.5 \pm 2\%$ for each culture and an impressive 98\% PDS Entropy. These findings highlight the effectiveness of using a MAS to create a globally pluralistic educational AI without the need for massive GPU clusters to train or fine-tune LLMs to achieve this level of cultural inclusivity. Although fine-tuning or training may be a more optimal approach, our methodology demonstrates that, with precise execution, these resource-intensive requirements can be circumvented.

\begin{figure}[t!]
    \centering
    \begin{subfigure}[t]{\columnwidth}
        \centering
        \includegraphics[width=0.8\columnwidth]{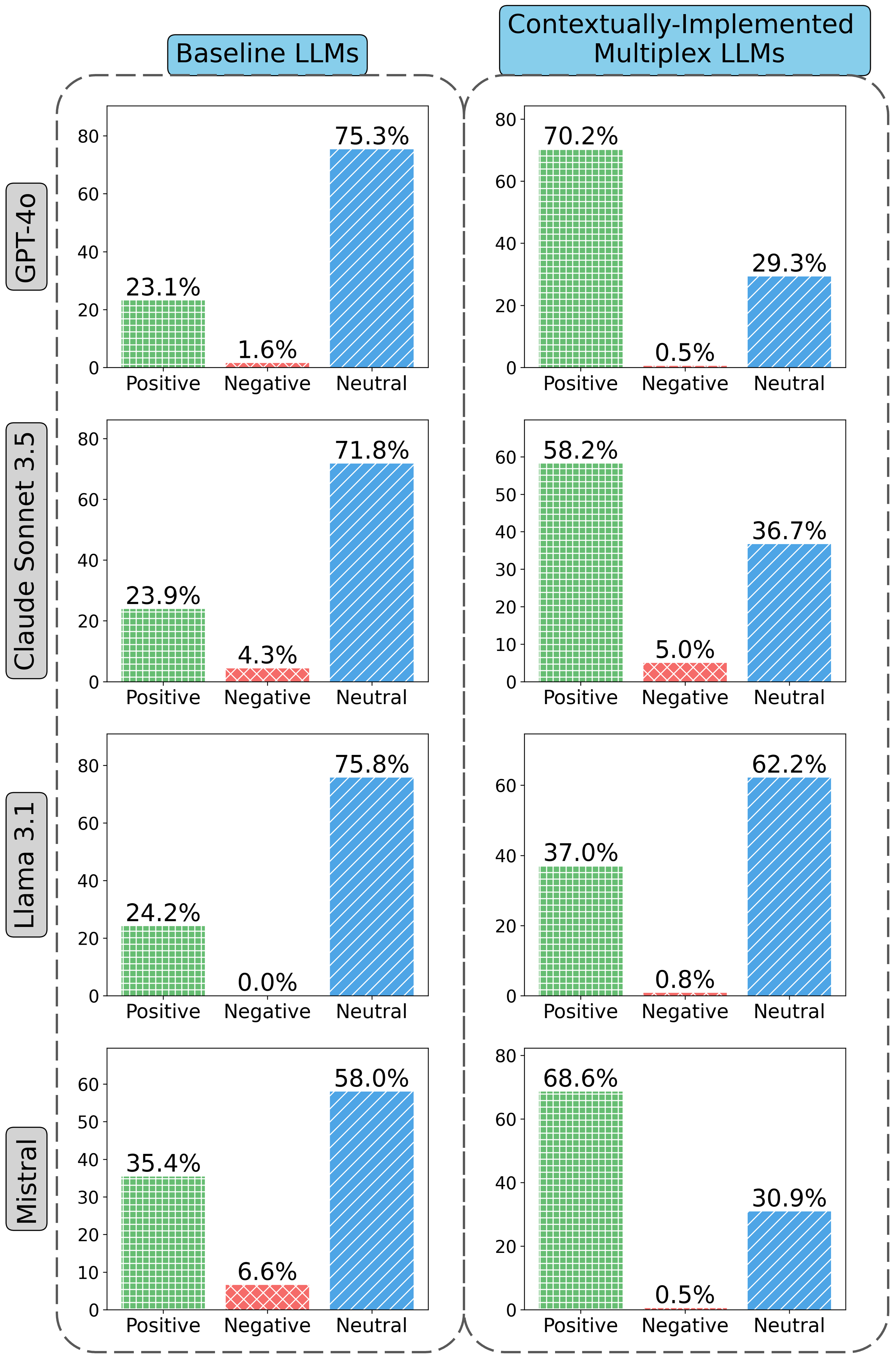}
        \caption{Sentiment Analysis for baseline and contextually-implemented multiplex LLM.}
        \label{fig:SentimentAnalysis_1_2}
    \end{subfigure}
    \begin{subfigure}[t]{0.47\columnwidth}
        \centering
        \includegraphics[width=\linewidth]{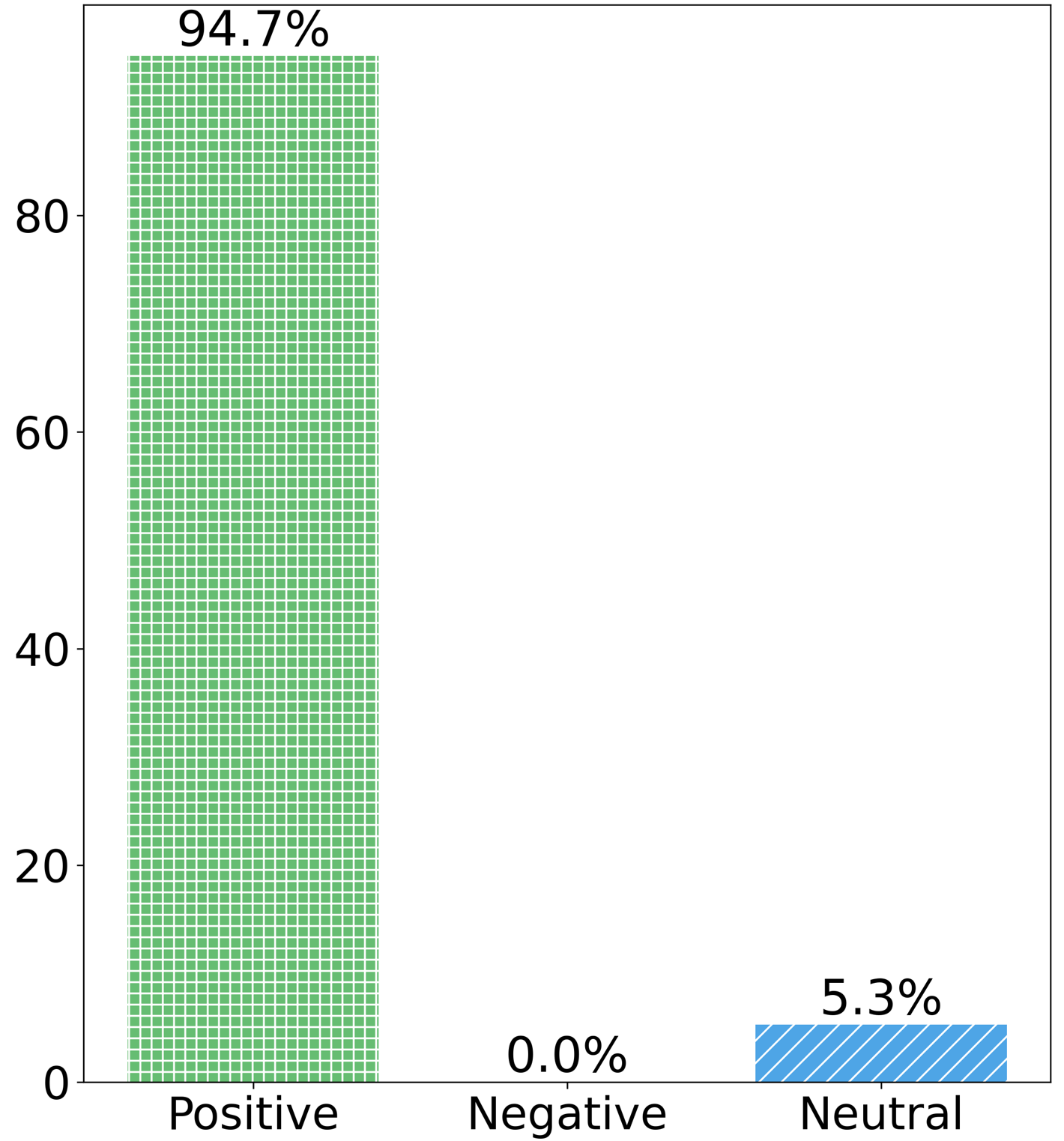}
        \captionsetup{width=\columnwidth}
        \caption{Sentiment Analysis for MAS-implemented (with GPT-4o) multiplex LLM. }
        \label{fig:SentimentAnalysis_3}
    \end{subfigure}
    \vspace{-2.5mm}
    \caption{Sentiment Analysis for the baseline LLM, and the two multiplex adaptations of these LLMs: the contextually-implemented multiplex models and the MAS-implemented multiplex models. \textit{A higher proportion of positive sentiment reflects the strategy's supportive stance across cultures, enhancing its effectiveness. Multiplex LLMs outperform baseline models, with the MAS-implemented version achieving the highest accuracy in sentiment analysis.}}
    \vspace{-6.5mm}
    \label{fig:SentimentAnalysis_combined}
\end{figure}

\vspace{-3mm}
\subsection{RQ3: Making LLMs Culturally Inclusive}
As discussed in \S\ref{subsec:Sentiment_Analyzer}, sentiment analysis is crucial to understanding how LLMs treat each culture, as helps us reveal the underlying biases within these models. Figure \ref{fig:SentimentAnalysis_1_2} illustrates the sentimental bias of these LLMs toward each culture, showing how different proposed mitigation strategies can make the LLMs more culturally appreciative. In this context, a green bar shows positive sentiment which indicates that the LLM is highly appreciative of cultures, a red bar shows negative sentiment suggesting opposition to that culture, and a blue bar shows neutral sentiment which generally implies that the model neither supports nor opposes the culture. However, as discussed in \S\ref{subsec:Sentiment_Analyzer} another perspective on neutral sentiment is that it reflects a lack of preference, displaying a non-appreciative tone toward these cultures.

In the \textit{Baseline Assessment} section of Fig \ref{fig:SentimentAnalysis_1_2}, we observe that the responses of LLMs are predominantly neutral rather than exhibiting a more positive (appreciative) stance. Some responses even reflect negative sentiment, indicating a considerable degree of bias. This baseline assessment highlights the inherent response style of each model.  In the \textit{Contextually Implemented Multiplexity} assessment section, on the other hand, we notice a shift from neutral and negative sentiments toward positive sentiments, indicating a broader appreciation of multiple cultures by the LLMs. Finally, Figure \ref{fig:SentimentAnalysis_3} illustrates that MAS-Implemented Multiplexity shows a clear, dominant shift from neutral and negative values to positive values. This demonstrates that the MAS responses are multicultural and highly appreciative toward all cultures, establishing it as the most effective strategy for achieving a globally pluralistic, educational AI.

\section{Conclusions}
\label{sec:Conclusions}

We propose a framework to establish a baseline for assessing cultural biases in Large Language Models (LLMs), defining cultural bias as a model's tendency to favor certain cultures over others. By analyzing LLM responses to educational questions, we find that these models often exhibit cultural polarization, with biases evident in both explicit content and subtle contextual cues. To address these biases, we introduce two strategies: (1) \textit{Contextually-Implemented} LLMs, which embed diverse cultural perspectives within the model's prompts to foster more inclusive responses, and (2) \textit{MAS-Implemented Multiplex} LLMs, where agents representing various cultural viewpoints contribute insights that are synthesized into a balanced response. Our results show that as mitigation strategies advance---from contextual prompting to MAS implementation---the Perspectives Distribution Score (PDS) improves, reflecting enhanced cultural inclusivity. The PDS Entropy score measures how evenly cultural perspectives are represented, with high entropy indicating balanced diversity. Baseline evaluations reveal a low average PDS Entropy of 3.25\%, which increases to 19\% with Contextually-Implemented LLMs and reaches 98\% with MAS-Implemented Multiplex LLMs, underscoring the effectiveness of these strategies. Sentiment analysis further reveals a shift from negative or neutral sentiment toward positive, with MAS-Implemented Multiplex LLMs achieving 0\% negative sentiment, highlighting improved cultural sensitivity. This study represents a pioneering effort to enhance cultural inclusivity in LLMs through the Multiplexity framework, establishing a baseline methodology and introducing practical mitigation strategies for fostering multicultural responses. We hope this work inspires further development of methodologies that expand the Multiplexity framework, advancing a truly global, pluralistic AI for education that respects and represents diverse cultural perspectives.

\bibliographystyle{IEEEtran}
\bibliography{bib}

\end{document}